\documentstyle[colacl]{article}
\title{{\bf Noun Phrase Recognition by System Combination}}
\author{
\begin{tabular}{cc}
{\bf Erik F. Tjong Kim Sang}\\
Center for Dutch Language and Speech\\
University of Antwerp\\
{\it erikt@uia.ua.ac.be}
\end{tabular}
}

\date{\today}

\begin{document}
\maketitle
\begin{abstract}
\noindent
The performance of machine learning algorithms can be improved by combining
the output of different systems.
In this paper we apply this idea to the recognition of noun phrases.
We generate different classifiers by using different representations
of the data.
By combining the results with voting techniques described in
\cite{vanhalteren98} we manage to improve the best reported 
performances on standard data sets for base noun phrases and
arbitrary noun phrases.
\end{abstract}

\section{Introduction}

\cite{vanhalteren98} and \cite{brill98} describe a series of successful
experiments for improving the performance of part-of-speech taggers.
Their results have been obtained by combining the output of different 
taggers with system combination techniques such as majority voting.
This approach cancels errors that are made by the minority of the
taggers.
With the best voting technique, the combined results decrease the 
lowest error rate of the component taggers by as much as 19\% 
\cite{vanhalteren98}.
The fact that combination of classifiers leads to improved performance
has been reported in a large body of machine learning work.

We would like to know what improvement combination techniques
would cause in noun phrase recognition.
For this purpose, we apply a single memory-based learning
technique to data that has been represented in different ways.
We compare various combination techniques on a part of the Penn
Treebank and use the best method on standard data sets for base 
noun phrase recognition and arbitrary noun phrase recognition.

\section{Methods and experiments}

In this section we start with a description of our task:
recognizing noun phrases.
After this we introduce the different data representations we use and
our machine learning algorithms.
We conclude with an outline of techniques for combining classifier
results.

\subsection{Task description}
\label{sec-task}

Noun phrase recognition can be divided in
two tasks: recognizing base noun phrases and recognizing arbitrary 
noun phrases.
Base noun phrases (baseNPs) are noun phrases which do not contain
another noun phrase.
For example, the sentence 

\begin{quote}
In $[$ early trading $]$ in $[$ Hong Kong $]$\\
$[$ Monday $]$ , $[$ gold $]$ was quoted at\\
$[$ \$ 366.50 $]$ $[$ an ounce $]$ .
\end{quote}

\noindent
contains six baseNPs (marked as phrases between square brackets).
The phrase {\it \$ 366.50 an ounce} is a noun phrase as well.
However, it is not a baseNP since it contains two other noun phrases. 
Two baseNP data sets have been put forward by \cite{ramshaw95}.
The main data set consists of four sections (15-18) of the Wall Street
Journal (WSJ) part of the Penn Treebank \cite{marcus93} as training
material and one section (20) as test
material\footnote{
 This \cite{ramshaw95} baseNP data set is available via
 ftp://ftp.cis.upenn.edu/pub/chunker/
}. 
The baseNPs in this data are slightly different from the ones that can
be derived from the Treebank, most notably in the attachment of
genitive markers.

The recognition task involving arbitrary noun phrases attempts to find
both baseNPs and noun phrases that contain other noun phrases.
A standard data set for this task was put forward at the CoNLL-99
workshop.
It consists of the same parts of the Penn Treebank as the
main baseNP data set: WSJ sections 15-18 as training data and section
20 as test data\footnote{
 Software for generating the data is available
 from http://lcg-www.uia.ac.be/conll99/npb/
}.
The noun phrases in this data set are the same as in the Treebank and
therefore the baseNPs in this data set are slightly different from the
ones in the \cite{ramshaw95} data sets.

In both tasks, performance is measured with three scores.
First, with the percentage of detected noun phrases that are correct
(precision). 
Second, with the percentage of noun phrases in the data that were
found by the classifier (recall).
And third, with the F$_{\beta=1}$ rate which is equal to
(2*precision*recall)/(precision+recall).
The latter rate has been used as the target for optimization.

\subsection{Data representation}
\label{sec-data-repr}

In our example sentence in section \ref{sec-task}, noun phrases are
represented by bracket structures.
Both \cite{munoz99} and \cite{tks99} have shown how classifiers can
process bracket structures.
One classifier can be trained to recognize open brackets (O) while
another will process close brackets (C).
Their results can be converted to baseNPs by making pairs of open and
close brackets with large probability scores \cite{munoz99} or by
regarding only the shortest phrases between open and close brackets as
baseNPs \cite{tks99}.
We have used the bracket representation (O+C) in combination with the
second baseNP construction method.

An alternative representation for baseNPs has been put forward by
\cite{ramshaw95}.
They have defined baseNP recognition as a tagging task:
words can be inside a baseNP (I) or outside of baseNPs (O).
In the case that one baseNP immediately follows another baseNP,
the first word in the second baseNP receives tag B.
Example:

\begin{quote}
In$_O$
early$_I$
trading$_I$
in$_O$
Hong$_I$
Kong$_I$
Monday$_B$
,$_O$
gold$_I$
was$_O$
quoted$_O$
at$_O$
\$$_I$
366.50$_I$
an$_B$
ounce$_I$
.$_O$
\end{quote}

\noindent
This set of three tags is sufficient for encoding baseNP structures
since these structures are nonrecursive and nonoverlapping.

\cite{tks99} have presented three variants of this tagging
representation.
First, the B tag can be used for the first word of every noun phrase
(IOB2 representation).
Second, instead of the B tag an E tag can be used to mark the last
word of a baseNP immediately before another baseNP (IOE1).
And third, the E tag can be used for every noun phrase final word
(IOE2).
They have used the \cite{ramshaw95} representation as well (IOB1).
We will use these four tagging representations as well as the O+C
representation. 

\subsection{Machine learning algorithms}

We have used the memory-based learning algorithm {\sc ib1-ig} which is part
of TiMBL package \cite{timbl99}.
In memory-based learning the training data is stored and a new item
is classified by the most frequent classification among
training items which are closest to this new item.
Data items are represented as sets of feature-value pairs.
In {\sc ib1-ig} each feature receives a weight which is based on the
amount of information which it provides for computing the
classification of the items in the training data.
These feature weights are used for computing the distance between a
pair of data items \cite{timbl99}.
{\sc ib1-ig} has been used successfully on a large variety of natural
language processing tasks.

Beside {\sc ib1-ig}, we have used {\sc igtree} in the combination
experiments. 
{\sc igtree} is a decision tree variant of {\sc ib1-ig}
\cite{timbl99}.
It uses the same feature weight method as {\sc ib1-ig}.
Data items are stored in a tree with the most important features close
to the root node.
A new item is classified by traveling down from the root node until a
leaf node is reached or no branch is available for the current feature
value.
The most frequent classification of the current node will be chosen.

\subsection{Combination techniques}
\label{sec-combi}

Our experiments will result in different classifications of the
data and we need to find out how to combine these.
For this purpose we have evaluated different voting mechanisms,
effectively the voting methods as described in \cite{vanhalteren98}.
All combination methods assign some weight to the results of the
individual classifier.
For each input token, they pick the classification score with the 
highest total score.
For example, if five classifiers have weights 0.9, 0.4, 0.8, 0.6 and
0.6 respectively and they classify some token as 
npstart,
null,
npstart,
null
and null,
then the combination method will pick npstart since
it has a higher total score (1.7) than null (1.6).
The values of the weights are usually estimated by processing a part
of the training data, the tuning data, which has been kept separate as
training data for the combination process.

In the first voting method, each of the five classifiers receives the
same weight (majority).
The second method regards as the weight of each individual
classification algorithm its accuracy on the tuning data
(TotPrecision).
The third voting method computes the precision of each assigned tag
per classifier and uses this value as a weight for the classifier in
those cases that it chooses the tag (TagPrecision).
The fourth method uses the tag precision weights as well but it
subtracts from them the recall values of the competing classifier
results.
Finally, the fifth method uses not only a weight for the current
classification but it also computes weights for other possible 
classifications.
The other classifications are determined by examining the tuning data
and registering the correct values for every pair of classifier results 
(pair-wise voting).

Apart from these five voting methods we have also processed the 
output streams with two classifiers: {\sc ib1-ig} (memory-based) 
and {\sc igtree} (decision tree).
This approach is called classifier stacking.
Like \cite{vanhalteren98}, we have used different input versions:
one containing only the classifier output and another 
containing both classifier output and a compressed representation 
of the classifier input.
For the latter purpose we have used the part-of-speech tag of the
current word.

\section{Results}

Our first goal was to find out whether system combination could
improve performance of baseNP recognition and, if this was the fact,
to select the best combination technique.
For this purpose we performed a 10-fold cross validation experiment on
the baseNP training data, sections 15-18 of the WSJ part of the Penn
Treebank (211727 tokens).
Like the data used by \cite{ramshaw95}, this data was retagged by the
Brill tagger in order to obtain realistic part-of-speech (POS) 
tags\footnote{
 No perfect Penn Treebank POS tags will be available for novel texts.
 If we would have used the Treebank POS tags for NP recognition, 
 our performance rates would have been unrealistically high.
}.
The data was segmented into baseNP parts and non-baseNP parts in a 
similar fashion as the data used by \cite{ramshaw95}.

The data was converted to the five data representations (IOB1, IOB2, IOE1,
IOE2 and O+C) and {\sc ib1-ig} was used to classify it by using
10-fold cross validation.
This means that the data was divided in ten consecutive parts of about
the same size after which each part was used as test data with the
other nine parts as training data.
The standard parameters of {\sc ib1-ig} have been used except for k,
the number of examined nearest neighbors, which was set to three.
Each word in the data was represented by itself and its POS tag and 
additionally a left and right context of four word-POS tag pairs.
For the first four representations, we have used a second processing
stage as well.
In this stage, a word was represented by itself, its POS tag, a
left and right context of three word-POS tag pairs and a left and
right context of  two classification results of the first processing
stage (see figure \ref{fig-ex}).
The second processing stage improved the F$_{\beta=1}$ scores with
almost 0.7 on average.

\begin{table}
\begin{center}
\begin{tabular}{|l|r|r|}\cline{2-3}
\multicolumn{1}{l|}{train} &
\multicolumn{1}{c|}{O} & \multicolumn{1}{c|}{C}\\\hline
All correct      & 96.21\% & 96.66\%\\
Majority correct &  1.98\% &  1.64\%\\
Minority correct &  0.88\% &  0.75\%\\
All wrong        &  0.93\% &  0.95\%\\\hline
\end{tabular}
\end{center}
\caption{Token classification agreement between the five classifiers
applied to the baseNP training data after conversion to the open
bracket (O) and the close bracket representation (C).}
\label{tab-res0}
\end{table}

The classifications of the IOB1, IOB2, IOE1 and IOE2
representations were converted to the open bracket (O) and close
bracket (C) representations. 
After this conversion step we had five O results and five C results.
In the bracket representations, tokens can be classified as either
being the first token of an NP (or the last in the C representation)
or not.
The results obtained with these representations have been measured
with accuracy rates: the percentage of tokens that were classified
correctly.
Only about one in four tokens are at a baseNP boundary so guessing
that a text does not contains baseNPs will already give us an accuracy
of 75\%. 
Therefore the accuracy rates obtained with these representations are
high and the room for improvement is small (see table \ref{tab-res0}).
However, because of the different treatment of neighboring chunks,
the five classifiers disagree in about 2.5\% of the classifications.
It seems useful to use combination methods for finding the best
classification for those ambiguous cases.

\begin{table}
\begin{center}
\begin{tabular}{|l|c|c|}\cline{2-3}
\multicolumn{1}{l|}{train} & O & C \\\hline
{\bf Representation} &&\\
IOB1             & 98.01\% & 98.14\% \\
IOB2             & 97.80\% & 98.08\% \\
IOE1             & 97.97\% & 98.04\% \\
IOE2             & 97.89\% & 98.08\% \\
O+C              & 97.92\% & 98.13\% \\\hline\hline
{\bf Simple Voting} &&\\
Majority         & 98.19\% & 98.30\% \\
TotPrecision     & 98.19\% & 98.30\% \\
TagPrecision     & 98.19\% & 98.30\% \\
Precision-Recall & 98.19\% & 98.30\% \\\hline
{\bf Pairwise Voting} &&\\
TagPair          & 98.19\% & 98.30\% \\\hline
{\bf Memory-Based} &&\\
Tags             & 98.19\% & 98.34\% \\
Tags + POS       & 98.19\% & 98.35\% \\\hline
{\bf Decision Trees} &&\\
Tags             & 98.17\% & 98.34\% \\        % igtree
Tags + POS       & 98.17\% & 98.34\% \\\hline  % igtree (lcgvote/text/logs)
\end{tabular}
\end{center}
\caption{Open and close bracket accuracies for the baseNP training
data (211727 tokens).
Each combination performs significantly better than any of the five
individual classifiers listed under Representation. 
The performance differences between the combination methods
are not significant.
}
\label{tab-res1}
\end{table}

\tabcolsep=0.08cm

\begin{figure*}
\begin{center}
\begin{tabular}{ccccccccc}
trading/NN &
in/IN &
Hong/NNP &
Kong/NNP &
\underline{Monday}/NNP &
,/, &
gold/NN &
was/VBD &
quoted/VBN \\
 &
in/IN &
Hong/NNP/I &
Kong/NNP/I &
\underline{Monday}/NNP &
,/,/O &
gold/NN/I &
was/VBD & \\
\end{tabular}
\end{center}
\caption{
Example of the classifier input features used for classifying {\it
Monday} in the example sentence.
The first processing stage (top) contains a word and POS context of
four left and four right while the second processing stage (bottom)
contains a word and POS context of three and a chunk tag context of
two. 
}
\label{fig-ex}
\end{figure*}

\tabcolsep=0.22cm

\begin{table*}
\begin{center}
\begin{tabular}{|l|c|c|c|c|}\cline{2-5}
\multicolumn{1}{l|}{section 20}
                 & accuracy & precision & recall & F$_{\beta=1}$\\\hline
Majority voting  & O:98.10\% C:98.29\% & 93.63\% & 92.89\% & 93.26 \\\hline
\cite{munoz99}   & O:98.1\%  C:98.2\%  & 92.4\%  & 93.1\%  & 92.8  \\
\cite{tks99}     & 97.58\%             & 92.50\% & 92.25\% & 92.37 \\
\cite{ramshaw95} & 97.37\%             & 91.80\% & 92.27\% & 92.03 \\
\cite{argamon98} & -                   & 91.6\%  & 91.6\%  & 91.6  \\\hline
\end{tabular}

\vspace*{0.7cm}
\begin{tabular}{|l|c|c|c|c|}\cline{2-5}
\multicolumn{1}{l|}{section 00}
                 & accuracy & precision & recall & F$_{\beta=1}$\\\hline
Majority voting  & O:98.59\% C:98.65\% & 95.04\% & 94.75\% & 94.90 \\\hline
\cite{tks99}     & 98.04\%             & 93.71\% & 93.90\% & 93.81 \\
\cite{ramshaw95} & 97.8\%              & 93.1\%  & 93.5\%  & 93.3 \\\hline
\end{tabular}
\end{center}
\caption{The results of majority voting of different data representations 
applied to the two standard data sets put forward by \cite{ramshaw95}
compared with earlier work.
The accuracy scores indicate how often a word was classified correctly
with the representation used (O, C or IOB1).
The training data for WSJ section 20 contained 211727 tokens while
section 00 was processed with 950028 tokens of training data.
Majority voting outperforms all earlier reported results for the two
data sets.} 
\label{tab-res2}
\end{table*}

The five O results and the five C results were processed by the
combination techniques described in section \ref{sec-combi}.
The accuracies per input token for the combinations can be found in
table \ref{tab-res1}. 
For both data representations, all combinations perform significantly
better than the best individual classifier (p$<$0.001 according to a
$\chi^2$ test)\footnote{ 
 We have performed significance computations on the bracket accuracy
 rates because we have been unable to find a satisfactory method
 for computing significance scores for F$_{\beta=1}$ rates.
}.
Unlike in \cite{vanhalteren98}, the best voting technique 
did not outperform the best stacked classifier.
Furthermore the performance differences between the combination
methods are not significant (p$>$0.05).
To our surprise the five voting techniques performed the same.
We assume that this has happened because the accuracies of the
individual classifiers do not differ much and because the
classification involves a binary choice.

Since there is no significant difference between the combination
methods, we can use any of them in the remaining experiments.
We have chosen to use majority voting because it does not require 
tuning data.
We have applied it to the two data sets mentioned in \cite{ramshaw95}.
The first data set uses WSJ sections 15-18 as training data 
(211727 tokens) and section 20 as test data (47377 tokens).
The second one uses sections 02-21 of the same corpus as training data
(950028 tokens) and section 00 as test data (46451 tokens).
All data sets were processed in the same way as described earlier.
The results of these experiments can be found in table \ref{tab-res2}.
With section 20 as test set, we managed to reduce the error of the best
result known to us with 6\% with the error rate dropping from 7.2\%
to 6.74\%,
and for section 00 this difference was almost 18\% with the error rate
dropping from 6.19\% to 5.10\%
(see table \ref{tab-res2}).

We have also applied majority voting to the NP data set put forward on 
the CoNLL-99 workshop.
In this task the goal is to recognize all NPs.
We have approached this as repeated baseNP recognition.
A first stage detects the baseNPs.
The recognized NPs are replaced by their presumed head word with a
special POS tag and the result is send to a second stage which
recognizes NPs with one level of embedding.
The output of this stage is sent to a third stage and this stage
finds NPs with two levels of embedding and so on.

In the first processing stage we have used the five data
representations with majority voting.
This approach did not work as well for other stages.
The O+C representation outperformed the other four representations by
a large margin for the validation data\footnote{
 The validation data is the test set we have used for estimating the
 best parameters for the CoNLL experiment: WSJ section 21.
}.
This caused the combined output of all five representations being
worse than the O+C result.
Therefore we have only used the O+C representation for recognizing 
non-baseNPs.
The overall system reached an F$_{\beta=1}$ score of 83.79 and this 
is slightly better than the best rate reported at the CoNLL-99 
workshop (82.98 \cite{conll99}, an error reduction of 5\%).

\section{Related work}

\cite{abney91} has proposed to approach parsing by starting with
finding correlated chunks of words.
The chunks can be combined to trees by a second processing stage,
the attacher.
\cite{ramshaw95} have build a chunker by applying
transformation-based learning to sections of the Penn Treebank.
Rather than working with bracket structures, they have represented
the chunking task as a tagging problem.
POS-like tags were used to account for the fact that words were inside
or outside chunks.
They have applied their method to two segments of the Penn Treebank
and these are still being used as benchmark data sets.

Several groups have continued working with the Ramshaw and Marcus data
sets for base noun phrases.
\cite{argamon98} use Memory-Based Sequence Learning for recognizing
both NP chunks and VP chunks.
This method records POS tag sequences which contain chunk boundaries
and uses these sequences to classify the test data.
Its performance is somewhat worse than that of Ramshaw and Marcus
(F$_{\beta=1}$=91.6 vs. 92.0) but it is the best result obtained
without using lexical information\footnote{
 We have applied majority voting of five data representations to
 the Ramshaw and Marcus data set without using lexical information
 and the results were:
 accuracy O: 97.60\%, accuracy C: 98.10\%, precision: 92.19\%, 
 recall: 91.53\% and F$_{\beta=1}$: 91.86.
}.
\cite{cardie98} store POS tag sequences that make up complete chunks
and use these sequences as rules for classifying unseen data. 
This approach performs worse than the method of Argamon et al.
(F$_{\beta=1}$=90.9).

Three papers mention having used the memory-based learning method 
{\sc ib1-ig}.
\cite{veenstra98} introduced cascaded chunking, a two-stage process
in which the first stage classifications are used to improve the
performance in a second processing stage.
This approach reaches the same performance level as Argamon et al. but
it requires lexical information.
\cite{daelemans99} report a good performance for baseNP recognition but
they use a different data set and do not mention precision and recall
rates.
\cite{tks99} compare different data representations for this task.
Their baseNP results are slightly better than those of Ramshaw and
Marcus (F$_{\beta=1}$=92.37).

\cite{xtag98} describes a baseNP chunker built from training data by a
technique called supertagging.
The performance of the chunker was an improvement of the Ramshaw and
Marcus results (F$_{\beta=1}$=92.4).
\cite{munoz99} use SNoW, a network of linear units, for recognizing
baseNP phrases and SV phrases.
They compare two data representations and report that a representation
with bracket structures outperforms the IOB tagging representation 
introduced by \cite{ramshaw95}.
SNoW reaches the best performance on this task (F$_{\beta=1}$=92.8).

There has been less work on identifying general noun phrases than on
recognizing baseNPs.
\cite{osborne99} extended a definite clause grammar with rules induced
by a learner that was based upon the maximum description length
principle.
He processed other parts of the Penn Treebank than we with an
F$_{\beta=1}$ rate of about 60.
Our earlier effort to process the CoNLL data set was performed in the 
same way as described in this paper but without using the combination
method for baseNPs.
We obtained an F$_{\beta=1}$ rate of 82.98 \cite{conll99}.

\section{Concluding remarks}

We have put forward a method for recognizing noun phrases by combining
the results of a memory-based classifier applied to different
representations of the data.
We have examined different combination techniques and each of them 
performed significantly better than the best individual classifier.
We have chosen to work with majority voting because it does not
require tuning data and thus enables the individual classifiers to use
all the training data.
This approach was applied to three standard data sets for base noun
phrase recognition and arbitrary noun phrase recognition.
For all data sets majority voting improved the best result for
that data set known to us.

Varying data representations is not the only way for generating different
classifiers for combination purposes.
We have also tried dividing the training data in partitions (bagging) and
working with artificial training data generated by a crossover-like
operator borrowed from genetic algorithm theory.
With our memory-based classifier applied to this data, we have been
unable to generate a combination which improved the performance of 
its best member.
Another approach would be to use different classification 
algorithms and combine the results.
We are working on this but we are still to overcome the practical 
problems which prevent us from obtaining acceptable results with the
other learning algorithms.

\section*{Acknowledgements}

We would like to thank
the members of the CNTS group in Antwerp, Belgium,
the members of the ILK group in Tilburg, The Netherlands and
three anonymous reviewers 
for valuable discussions and comments.
This research was funded by the European TMR network Learning
Computational Grammars\footnote{http://lcg-www.uia.ac.be/}.

\small
\bibliographystyle{acl}

\end{document}